\documentclass{article}
\usepackage{spconf,amsmath,graphicx,amssymb,bm}
\usepackage{textcomp}
\usepackage{cite}

\title{Exploring RNN-Transducer for Chinese speech recognition}
\name{Senmao Wang$^{1,3}$,Pan Zhou$^{2}$, Wei Chen$^{3}$, Jia Jia$^{2}$, Lei Xie$^{1}$
}
\address{$^{1}$ School of Computer Science,Northwestern Polytechnical University, Xi'an, China\\
$^{2}$Tiangong Institute for Intelligent Computing, Tsinghua University, Beijing, China
\\ $^{3}$Voice Interaction Technology Center, Sogou Inc., Beijing, China
\\\small \texttt{\{swang,lxie\}@nwpu-aslp.org,}\;\texttt{\{zh-pan,jjia\}@mail.tsinghua.edu.cn,}\;\texttt{chenweibj8871@sogou-inc.com}
}
\begin{document}
\ninept
\maketitle
\begin{abstract}
End-to-end approaches have drawn much attention recently for significantly simplifying the construction of an automatic speech recognition (ASR) system. RNN transducer (RNN-T) is one of the popular end-to-end methods. Previous studies have shown that RNN-T is difficult to train and a very complex training process is needed for a reasonable performance. In this paper, we explore RNN-T for a Chinese large vocabulary continuous speech recognition (LVCSR) task and aim to simplify the training process while maintaining performance. First, a new strategy of learning rate decay is proposed to  accelerate the model convergence. Second, we find that adding convolutional layers at the beginning of the network and using ordered data can discard the pre-training process of the encoder without loss of performance. Besides, we design experiments to find a balance among the usage of GPU memory, training circle and model performance. Finally, we achieve 16.9\% character error rate (CER) on our test set, which is 2\% absolute improvement from a strong BLSTM CE system with language model trained on the same text corpus.
\end{abstract}
\vspace{-10pt}
\section{Introduction}\vspace{-5pt}
\label{sec:inrto}
Most state-of-the-art automatic speech recognition (ASR) systems\cite{hinton2012deep, dahl2012context, sak2014long, sainath2015convolutional} have three main components, acoustic model, pronunciation model and language model. An unified model taking audio feature as input, acoustic model uses deep neural network (DNN) in combination with hidden Markov model (HMM)\cite{rabiner1989tutorial} and outputs posteriors of context dependence (CD) states. The decision tree based clustering connects context dependent states to phones. A separated expert-curated pronunciation model maps phones to words. Language model is used to construct words to a whole meaningful sentence. The three components are trained separately, and thus an end-to-end system\cite{graves2013generating} which combines all these components into one model attracts much of interest. Taking speech feature as input and outputting word symbols directly will remove the gap between different components. In other words, one model may avoid local optimum in three components and get global optimum. 

Compared with a traditional ASR system, an end-to-end ASR system aims to map the input speech sequence to the output graphme/word sequence using neural network. The main problem of this goal is that the length of input sequence and the length of output sequence are apparently different. 
To resolve this, encoder-decoder architecture is recommended. Combining with attention mechanism that aligns input to output, this architecture shows a good performance in various sequence-to-sequence mapping task, such as neural machine translation \cite{bahdanau2014neural}, text summarization \cite{rush2015neural}, image captioning \cite{xu2015show}, etc. Recently, attention based approaches\cite{bahdanau2016end, chorowski2014end, chorowski2015attention, shan2017attention} have been reported to perform well in ASR as well. The Listen, Attend and Spell (LAS) model,  proposed in \cite{chan2016listen}, uses a BLSTM network to map acoustic feature to a high level representation, and then uses an attention based decoder to predict output symbol on condition of previous output symbol. However, attention based decoder has to wait the formation of the entire high level representation for computing attention weights. It cannot be used in real-time tasks. 

The connectionist temporal classification (CTC) model\cite{graves2006connectionist,graves2014towards} can be regard as another kind of end-to-end approach. The blank label and the mapping function are used to align input sequence to target symbols. CTC based acoustic model usually use unidirectional or bidirectional LSTM as encoder and compute the CTC loss between the output sequence of encoder and the target symbol sequence. In \cite{sak2015fast}, context dependent phoneme (CDPh) based CTC model has shown good performance in an ASR system. In CTC based model, the frame independent assumption ignores the context information in some degree. Obviously, this assumption does not match the actual situation where content information is essential for speech modeling. Another disadvantage of CTC is that the length of the output posterior probability sequence must be longer than the length of the label sequence. This limitation does not allow for a large subsample rate for a CDPh based CTC system. 

RNN Transducer (RNN-T)~\cite{graves2012sequence,graves2013speech} has been recently proposed as an extension of the CTC model. Specifically by adding an LSTM based prediction network, RNN-T removes the conditional independence assumption in the CTC model. Moreover, RNN-T does not need the entire utterance level representation before decoding, which makes streaming end-to-end ASR possible. In\cite{rao2017exploring}, Google has implemented the RNN-T model to a streaming English ASR system and has achieved a comparative performance with conventional state-of-the-art speech recognition systems. As RNN-T is extended from an CTC acoustic model, it is usually initialized from a pre-trained CTC model. The hierarchical CTC (HCTC) architecture\cite{fernandez2007sequence} also can be used for a better initialization\cite{rao2017exploring}. Specifically in an English ASR task, they used phoneme CTC loss and grapheme CTC loss to assist wordpiece CTC loss optimization. The HCTC architecture helps to train a better model for initialization, which is beneficial to the RNN-T model.


\begin{figure}[t]
\centering
\includegraphics[width=5.5cm,height=5cm,angle=-0]{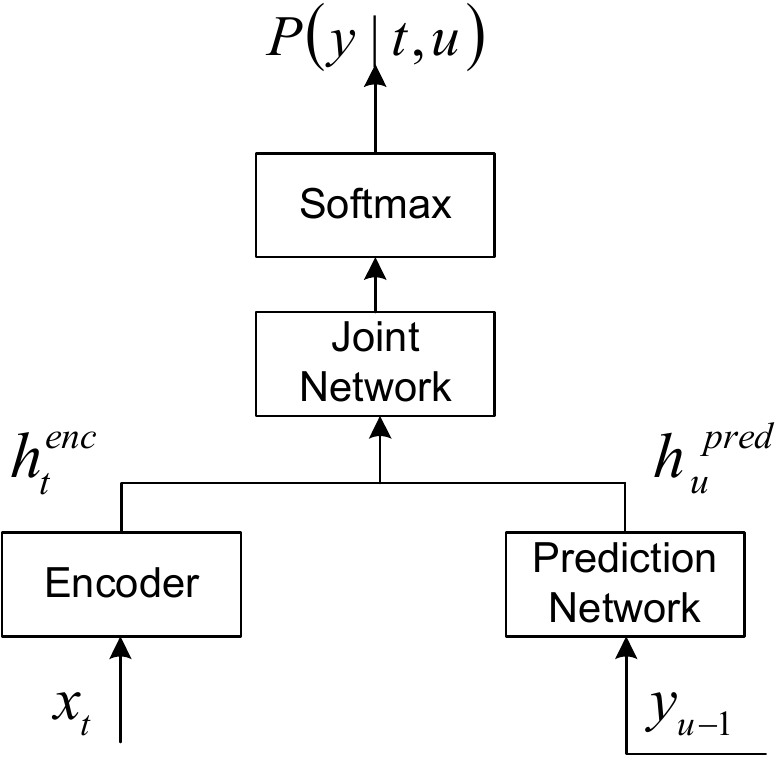}\vspace{-5pt}
\caption{Illustration of the RNN-Transducer model.}
\label{fig:1}\vspace{-20pt}
\end{figure}

In RNN-T training, for a good performance, it is necessary to use a pre-trained CTC model to initialize the encoder of the RNN-T model. And for a decent pre-trained CTC model, we usually use frame-wise cross entropy (CE) training to obtain a CE model as the start point of CTC training. Obviously, this long step-by-step process will cost lots of time in order to obtain a well-trained RNN-T model. 
In this paper, we explore the potentials of the RNN-T architecture on a Mandarin LVCSR task and attempt to simplify the training process. The main contributions of our work are as follows.
First, for a good model convergence, we propose an effective learning rate decay strategy which shapely decreases the learning rate at the epoch of the loss starting to increase and halves the learning rate every epoch after that. 
Second, we find that adding convolutional layers before the BLSTM layers can replace the functionality of a pre-trained CTC-based model for the encoder, leading to a simplified training process. 
Besides, to further accelerate the training process while ensuring performance, we compare the influence of different architectures and the subsampling rates in the encoder. We find that subsampling is a necessary step for accelerating training as the output of RNN-T is a four-dimension tensor which occupies too many GPU memories and slows down the training cycle. 
Furthermore, a pre-trained LSTM language model is proven to have a positive influence. Finally we achieve 16.9\% character error rate (CER) on our test set, which is 2\% absolute improvement from a strong BLSTM CE system with language model trained on the same text corpus.


\vspace{-10pt}
\section{Recurrent Neural Network-Transducer}
\label{sec:rnnt}
As mentioned earlier, RNN-T\cite{graves2012sequence} is an extension to the CTC~\cite{graves2006connectionist} model. On the base of the CTC encoder, Graves proposed to add an LSTM to learn context information, which functions as a language model. A joint network is subsequently used to combine the acoustic representation and the context representation together to compute posterior probability. Fig.~\ref{fig:1} illustrates the three main components of RNN-T, namely encoder, prediction network and joint network. 
Concretely, the encoder transforms the input time-frequency acoustic feature sequence $\boldsymbol{x}=(x_0,\dots,x_T)$ to a high-level feature representation $\boldsymbol{h}^{enc}$. 

\begin{equation}
\label{eq:encoder}
\bm{h}^{enc}=\text{Encoder}(\bm{x})
\end{equation}

Prediction network can remove the limitation of the frame independent assumption in the original CTC architecture. It usually adopts an LSTM to model context information which leads to transformation of the original one-hot vector $y=(y_1,\dots,y_U)$ to a high-level representation $h_u^{pred}$.
The output of the prediction network is determined by the previous context information. Note that the first input of the prediction network is an all-zero tensor and $y_{u-1}$ is the last non-blank unit. Eq. (\ref{eq:embeding}) and Eq. (\ref{eq:predictNet}) describe how the prediction network operates at label step $u$.
\begin{equation}
\label{eq:embeding}
h^{embed}_{u-1} =\text{Embedding}(y_{u-1})
\end{equation}
\begin{equation}
\label{eq:predictNet}
h^{pred}_{u} = \text{Prediction}(h^{embed}_{u-1})
\end{equation}
 $h_t^{enc}$ and $h_u^{pred}$ are first reshaped to $N\times T\times U\times H$ tensor where $N$ and $H$ represent  the batch size and the number of hidden nodes respectively. The joint network is usually a feed-forward network that produces $h_{t,u}^{joint}$ from $h_t^{enc}$ and $h_u^{pred}$. 
\begin{equation}
\label{eq:jointnet}
h_{t,u}^{joint}=tanh(\bm{W^{enc}}h_{t}^{enc} + \bm{W^{pred}}h_{u}^{pred} + b)
\end{equation}

Finally the output probability distribution is computed by a softmax layer:
\begin{equation}
\label{eq:softmax}
P(k|t,u)= \text{SoftmaxDist}(h_{t,u}^{joint})
\end{equation}
where $k$ is the index of the output classes.

Finally, the whole network is trained by optimizing the RNN-T loss which is computed by the forward-backward algorithm.
\begin{equation}
\label{eq:loss}
\bm{Loss}_{rnnt}=-ln\sum_{(t,u):t+u=n}p(y|t,u).
\end{equation}

In the decoding, the most likely sequence of characters is generated by the beam search algorithm. During the RNN-T inference, the input of the prediction network is the last non-blank symbol. The final output sequence is obtain by removing all blank symbols in the most likely sequence path. The temperature of the softmax function can be used to smooth the posterior probability distribution and benefit larger beam width. At the same time, an N-gram language model trained on external text can be integrated in the beam score. 

\vspace{-10pt}
\section{Dataset and Baseline}
\label{sec:exp}
We investigated the RNN-T model on a Chinese LVCSR task.  Specifically, we carried out a series of experiments in order to obtain good result on the Chinese task and to simplify the RNN-T model training at the same time.

\vspace{-10pt}
\subsection{Data}
\label{ssec:data}    
Our dataset is composed of approximately 1,000 hours of Mandarin speech collected by Sogou voice input method (IME) on mobile phones. We split out 50 hours from this dataset as validation and the rest 950 hours are used for training. 40-dimensional log Mel-filterbank coefficients are extracted from 25 ms frames shifted 10 ms. Global mean and variance normalization (CMVN) is performed to achieve our final acoustic feature. To obtain more convincing results, we use a rich test set which is recorded by Sogou IME in different clean and noisy environments. Each test set has around 8000 utterances and in total we have 17.4 hours for testing. In order to train a truly end-to-end model, we choose Chinese character as our modeling unit. We use a symbol inventory consisting of 26 English characters, 6784 frequently-used Chinese characters, an unknown token (UNK) and a blank token.

\vspace{-10pt}
\subsection{Baseline}
\label{ssec:baseline}

We train several models as our baselines. First, a 4-layer BLSTM model with 256 cells/layer is trained by optimizing frame-level CE loss using tied phone states as targets. Meanwhile, using the same architecture, we also train two CTC models using the monophone-level CTC loss and the character-level CTC loss.  We also train several RNN-T models using the standard pretraining method in~\cite{rao2017exploring}. Our base RNN-T shares the same encoder as the CTC model except the one uses HCTC pre-training method which has 5 layers of BLSTM. Following the work in~\cite{rao2017exploring, fernandez2007sequence}, for HCTC pre-training, we increase the number of  BLSTM layer to 5 and add a monophone-level CTC loss after the third layer. The prediction network is a 2-layer LSTM with 512 cells/layer.  One fully-connected feed-forward neural network with 512 nodes is used as the joint network. 

We concatenate 7 frames (3-1-3) of FBank features as the network input. Frame skipping (2 frames) is also adopted in the CTC and RNN-T models. Adam optimizer is used to learn parameters and the initial learning rate is 0.0002.

\begin{table}[t]
\centering
\caption{\textit{Performance of the baseline systems in CER.}}
\begin{tabular}{|l|l|}
\hline
Model & CER (\%) \\ \hline\hline
CE  & 18.87 \\ \hline
charCTC  & 20.93 \\ \hline
phoneCTC  & 19.06 \\ \hline \hline
RNN-T  & 22.39 \\ \hline
\ +CTC init & 20.83 \\ \hline
\ \ +LM init & 19.98 \\ \hline
\ \ \ +HCTC init  & \textbf{19.05} \\ \hline
\ \ \ \ +Beam 10 &  \textbf{18.78} \\ \hline
\end{tabular}
\label{tablebaseline}\vspace{-10pt}
\end{table}

The decoding process of the CTC based model follows the setup in~\cite{ miao2015eesen}. Unless elsewhere stated, The RNN-T decoding uses a beam-width of 5 without an external language model. We summarize our baseline results in Table~\ref{tablebaseline}. We can see that the two CTC models are worse than the CE state model. This is mainly because we use monophones as the modeling unit in the CTC model. We believe the performance will be better if we use context dependent phones as the modeling unit. As expected, the random initialized RNN-T model performs much worse than the CTC and CE models. It can be found that with proper encoder and prediction network initialization methods, RNN-T gradually improves from 22.39\% to 18.78\% in terms of CER, which eventually surpass the CE and CTC models. 

Although our baseline RNN-T model eventually performs better than the CE and CTC models, it is obvious that the training process is sophisticated, involving many pre-training initialization steps according to~\cite{rao2017exploring}. In the following, we will discuss the proposed tricks to simplify the training circle of an RNN-T model.

\vspace{-10pt}
\section{Proposed Tricks}
\subsection{Sharpen learning rate decay}
\label{ssec:sharpLearnRate}

\begin{figure}[t]
\centering
\includegraphics[height=5cm,width=7cm]{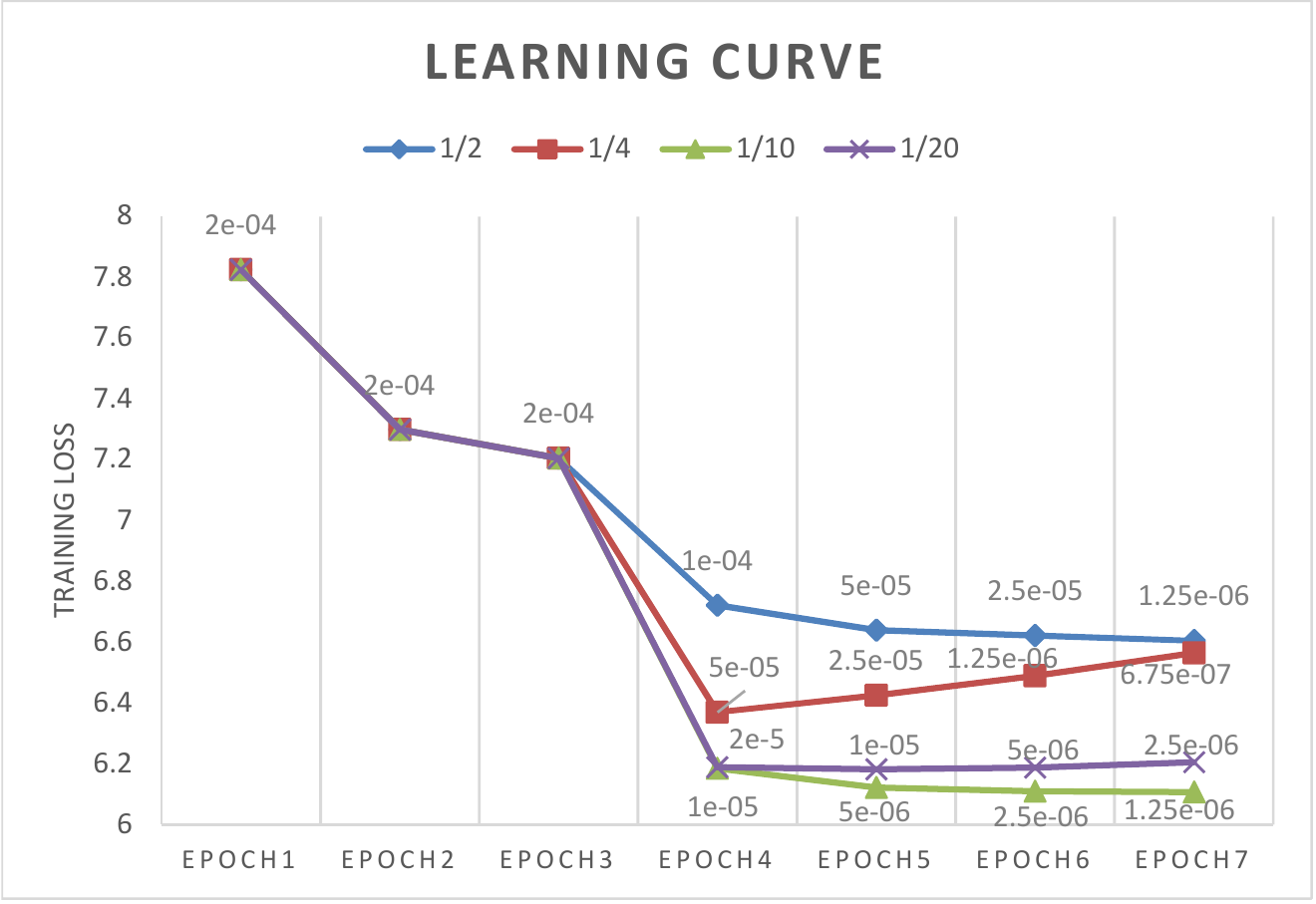}
\caption{Learning curves of RNN-T.}\vspace{-10pt}
\label{fig:learncurve}\vspace{-10pt}
\end{figure}

During training of the baseline RNN-T models, we find that this kind of model is hard to train and easy to get overfitted. The common setting in neural network training is as follows. The learning rate remains fixed for the first few epochs before the loss on the validation set begins to increase, and then it is divided by 2 every epoch after that.  In order to overcome the overfitting problem, we try a different learning rate decay strategy which brings clearly positive effect. Specifically, we use a more aggressive strategy which divides the learning rate by a number larger than 2 at the first decay epoch and it changes as usual in the following epochs. Fig.~\ref{fig:learncurve} illustrates how the training loss changes over training epochs for different learning rate decay strategy. A significant decline in the training loss is observed when the learning rate is first divided by more than 2. The best model coverage is achieved when the learning rate is first divided by 10. By applying this strategy, we achieve a clear improvement over the baseline, as shown in Table~\ref{tableLearnrate}.

Besides, dropout is also a common trick to cope with the overfitting problem. We find that with a dropout probability equals to 0.2, we improve our RNN-T model from 19.98\% to 19.51\% in terms of CER. In the following experiments, the sharpen learning rate decay strategy (1/10) and a dropout rate of 0.2 are adopted.

\begin{table}[t]
\centering
\caption{\textit{The impact of sharpen learning rate decay strategy and droput. Here RNN-T+LM init in Table~\ref{tablebaseline} is used as baseline and we denote it as RNN-T for simplicity. }}
\begin{tabular}{|l|l|}
\hline
Model config	&  CER  (\%) \\ \hline
RNN-T  & 19.98  \\ \hline
+sharpen  & 19.69 \\ \hline
+dropout &19.51 \\ \hline
\end{tabular}
\label{tableLearnrate}\vspace{-10pt}
\end{table}

\vspace{-5pt}
\subsection{Abandon encoder pretraining}
\label{ssec:addCNN}

As mentioned in Section~\ref{ssec:baseline}, we need to use a trained CTC model to initialize the encoder of the RNN-T model. Moreover, we also need to use CE training to initialize the CTC model. The training process is really complicated, which costs too much time. We wonder if the pre-training process is necessary for the RNN-T model and try to abandon this complicated training process.

First, as convolutional neural networks (CNN) can help to extract more invariant and stable features, we incorporate CNN layers in the encoder to explore how it affects the performance. Two layers of CNN with 6x6 kernel size are added before the BLSTM layer. Besides, curriculum learning (CL), which makes the model to learn from an easy task to a hard task, can also be cosidered to accelerate training convergence. We use CL by sorting the training sentences according to their length.

Table \ref{tableCNN} shows the effects of RNN-T with CNN layers and CL training. Here we select the RNN-T model in the second last row in Table \ref{tablebaseline} as the base model, renamed it as RNN-T (enc init), meaning the use of encoder initialization (HCTC). Note that we use dropout and sharpen learning decay rate here, so the CER of this model is lower than that in Table~\ref{tablebaseline}. We find that by adding two CNN layers, we can achieve a CER of 17.65\% from random initialization, with absolute 1\% CER reduction as compared with HCTC-initialization trained RNN-T. Note that CL also helps for a lower CER. But when the two strategies are used together, we obtain a small performance degradation. Hence in conclusion, the complicated RNN-T training process can be removed by adding CNN layers to the RNN-T model.

\begin{table}[t]
\centering
\caption{\textit{The impact of CNN layers and curriculum learning.}}
\begin{tabular}{|l|l|}
\hline
  	Model & CER (\%)\\ \hline\hline
RNN-T (enc init) &18.62 \\ \hline
\ +2 CNN layers (rand init)& \textbf{17.65} \\ \hline \hline
RNN-T (enc init) + CL &18.45 \\ \hline
\ +2 CNN layers (rand init)  &17.94 \\ \hline
\end{tabular}
\label{tableCNN}\vspace{-10pt}
\end{table}

\subsection{Acceleration by subsampling}
\label{ssec:subsample}
\begin{table}[t]
\centering
\caption{\textit{RNN-T performance of different subsampling configuration in encoder with 2 CNN layers and 5 BLSTM layers. MP2@1-2 represents maxpooling  with size=2 at 1st and 2nd CNN layer; Py2@1-3 means pyramid size=2 at 1st, 2nd and 3rd BLSTM layer.}}
\begin{tabular}{|c|c|c|}
\hline
total subsample &subsample config & CER (\%)\\ \hline
2 &	MP2@2  & 18.32 \\ \hline
2 &	Py2@3 & 18.39  \\ \hline \hline
4 &	MP2@2+Py2@3 & 18.07 \\ \hline
4 &	MP2@1-2 & 18.30 \\ \hline
4 &	Py2@1-2 & 18.69 \\ \hline
4& 	Py2@2-3 & \textbf{17.94} \\ \hline
4 & 	Py2@3-4 & 17.98 \\ \hline 
4 &  	Py2@4-5 & 18.11 \\ \hline \hline
6& 	MP2@2+Py3@3 & 17.95 \\ \hline \hline
8&	MP2@2+Py2@2-3 & 18.58\\ \hline
8&	MP2@1-2+P2@3 & 18.88\\ \hline
8&	Py2@1-3& 18.42\\ \hline
\end{tabular}
\label{tableSubsample}\vspace{-10pt}
\end{table}

The output of RNN-T is a four-dimension tensor representing thousands of label classes and hundreds of speech frames. Consequently, it will cost a large amount of GPU memory. Larger batch size is another way to accelerate the training speed. Although we use a NIVDIA M40 GPU with 24GB memory, we still cannot increase our batch size to a reasonable number under the configuration of skipping frame number 2. Higher skipping frame rate causes shorter acoustic features to save memory; however, performance will degrade when more frames are skipped. In the CNN and BLSTM equipped encoder, we try to exploit subsampling within the layer by max-pooling (MP) after the CNN layer and the pyramid BLSTM (pBLSTM). pBLSTM is a BLSTM layer whose input is obtained by concatenating several frames of its preceding layer outputs. Unless otherwise stated the max-pooling size is set to 2 and the pyramid size is set to 2 when pBLSTM takes 2 frames of its input features and skips 2 frames. 

In our RNN-T model, the encoder is composed of 2 CNN layers and 5 BLSTM layers. We compare different subsampling configurations, including size and location of the max-pooling and pyramid layer. Table~\ref{tableSubsample} shows the details of subsampling. It can be found that max-pooling and pyramid achieve similar performance at a total subsampling rate of 2. As the total subsampling rate is increased to 4 for a faster training speed, we find subsampling all in BLSTM parts is a better option than using max-pooling.  This maybe attribute to the information loss in max-pooling. For a subsampling rate larger than 6, performance degradation is observed. The most suitable subsampling rate is between 4 and 6 according to our results.

We choose the second and third layers as the pyramid BLSTM in our model. Using 24GB GPU memory, we can only set the batch size to 10 with the frame skipping rate of 1, but the batch size can be set as 20 with our subsampling ratio which accelerates the training.   

\begin{table}[t]
\centering
\caption{\textit{The impact of different initializations of prediction network.}}
\begin{tabular}{|l|l|}
\hline
Model config & CER \\ \hline
2-layer LSTM random init. & 17.94  \\ \hline
2-layer LSTM  init. w/ training transcription  & \textbf{17.61} \\ \hline
2-layer LSTM  init. w/ external text corpus   & 17.77 \\ \hline
\end{tabular}
\label{tableLMinit}
\end{table}

\begin{figure}[t]
\centering
\includegraphics[height=5cm,width=8cm]{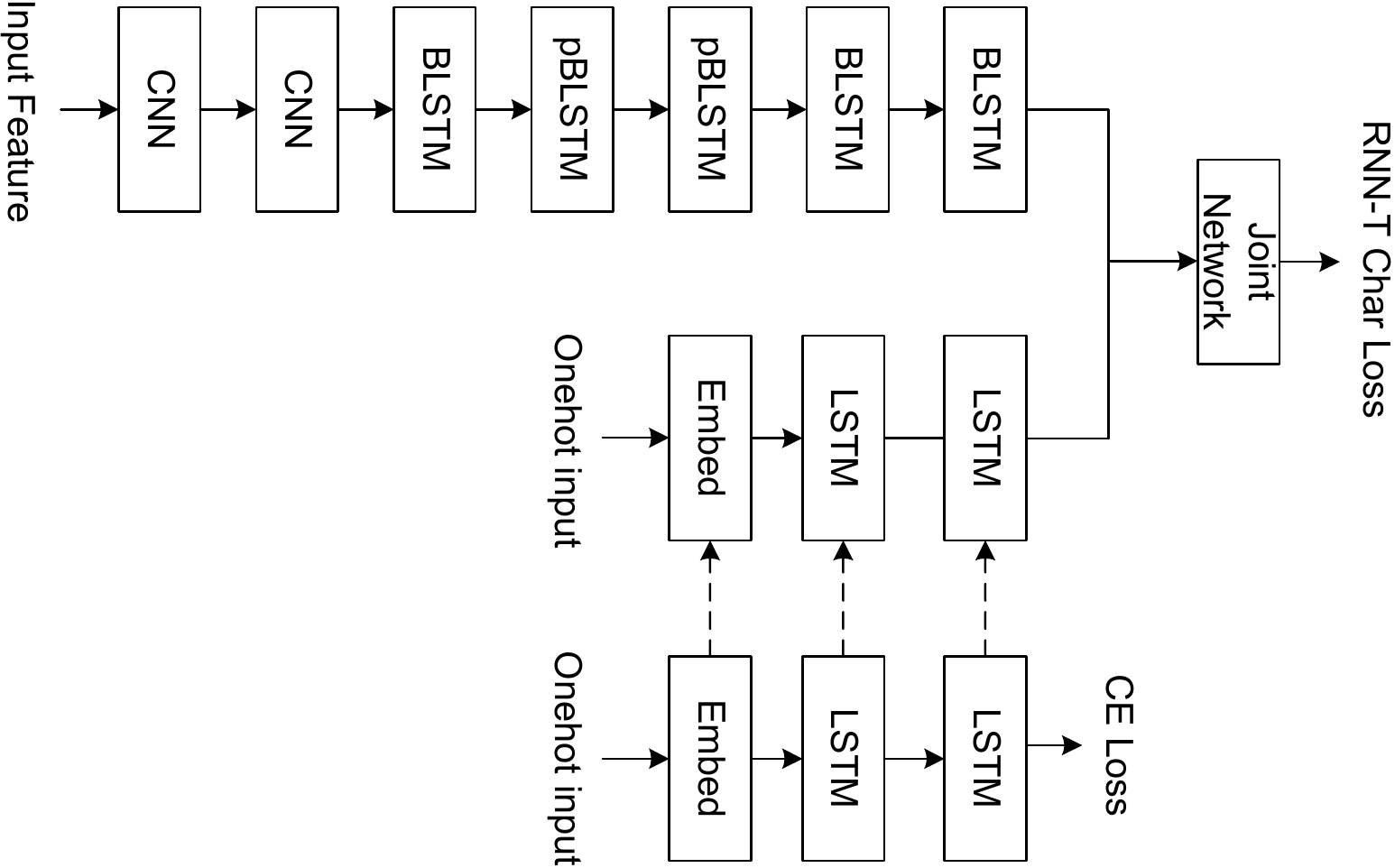}
\caption{The RNN-T architecture used in our paper, consisting of 2 layers of CNN and 5 layers of BLSTM, is trained from scratch. The prediction network is initialized by a character-level LSTM LM.}
\label{fig:2}
\end{figure}

\vspace{-5pt}
\subsection{Prediction network initialization}
\label{ssec:LMinit}

We further study the initialization strategy for the prediction network. Specifically, we train 2-layer LSTM language models using the training set transcriptions and an external 27G text corpus, respectively, and use them to initialize the training of the prediction network with the same LSTM structure. Here we use the best encoder architecture (total subsample 2, Py2@2-3 in Table~\ref{tableSubsample}) for the experiments, combining it with different prediction networks. Results are listed in Table~\ref{tableLMinit}. We find that initializing the prediction network with a same structure language model can bring performance improvement. The prediction network initialized by the LSTM language model trained using transcriptions from the training set itself shows better performance. It seems that the language model trained from the external corpus has domain mismatch with our experimental data.

\vspace{-5pt}
\subsection{Final results}
\label{combine}

We combine all of our previous mentioned tricks together and draw the final architecture of our RNN-T in Fig.~\ref{fig:2}. As listed in Table \ref{table1whr}, our best RNN-T model without encoder CTC-pretraining achieves a CER of 16.90\% with the help of an external 5-gram character LM, which is about 2\% absolute CER reduction from our BLSTM CE baseline. It also shows that the training transcriptions, corresponding to the 1000 hours of speech, are not enough for a decent language model in an end-to-end ASR system. 

We finally report the results on a task of Mandarin ASR using 10,000 hour of training data. Our RNN-T eventually achieves a CER of 10.52\%, without the use of an external language model. As a comparison, an internal Latency-Controlled BLSTM system has achieved a CER of 11.30\% on the same test set.

\begin{table}[h]
\centering
\caption{\textit{Performance of RNN-T model trained on 1000h and 10000h Mandarin corpus.}}
\begin{tabular}{|l|l|l|}
\hline
Task & Model config  & CER (\%)\\ \hline
1,000hr & RNN-T & 17.61 \\ \hline
 1,000hr & \ + Character LM & 16.90  \\ \hline
10,000hr & RNN-T & \textbf{10.52}\\ \hline
10,000hr & LC-BLSTM & 11.30 \\ \hline
\end{tabular}
\label{table1whr}\vspace{-10pt}
\end{table}

\vspace{-5pt}
\section{Conclusion}
In this work, we have explored the RNN-T model training in a Mandarin LVCSR task. We have proposed several methods to speed up RNN-T training, including sharpen learning rate decay strategy, abandon encoder pre-training by adding CNN layers, accelerating training by proper subsampling and LM initialization. Finally we achieve a simplified training procedure for RNN-T with a superior performance as compared to a strong BLSTM CE system.  
\clearpage
\bibliographystyle{IEEEbib}
\bibliography{wsm-2019}

\end{document}